\documentclass[onecolumn]{article}
\usepackage[utf8]{inputenc}
\usepackage[english]{babel}

\usepackage{geometry}
\usepackage{wrapfig}
\usepackage{times}
\usepackage{epsfig}
\usepackage{graphicx}
\usepackage{amsmath}
\usepackage{amssymb}
\usepackage{color}
\usepackage{algorithm}
\usepackage[noend]{algpseudocode}

\algnewcommand{\LeftComment}[1]{\Statex \textit{#1}}
\usepackage[font=footnotesize, caption=false]{subfig}
\usepackage[percent]{overpic}
\usepackage[export]{adjustbox}
\usepackage{mathtools}
\usepackage{tikzscale}
\usepackage{subfig}
\usepackage{savesym}

\usepackage{enumitem}
\usepackage{txfonts}
\usepackage{yfonts}
\usepackage{eufrak}
\usepackage{bbm}
\usepackage{dsfont}
\usepackage{enumitem}
\usepackage[output-decimal-marker={.}]{siunitx}

\usepackage[utf8]{inputenc}
\usepackage{tikz}
\usetikzlibrary{snakes,arrows,shapes}

\definecolor{blue(pigment)}{rgb}{0.2, 0.2, 0.6}
\definecolor{asparagus}{rgb}{0.53, 0.66, 0.42}

\usepackage{tikz}
\usetikzlibrary{snakes,arrows,shapes}

\usepackage{hyperref}

\vspace{-5ex}
\title{Using cameras for precise measurement of two-dimensional plant features: CASS}
\date{}

\author{Amy Tabb \\
	United States Department of Agriculture, Agricultural Research Service,\\
	Appalachian Fruit Research Station (USDA-ARS-AFRS)\\
	Kearneysville, West Virginia, USA\\
	{\tt\small amy.tabb@usda.gov}
	\and
	Germ\'{a}n A Holgu\'{i}n \\
	Electrical Engineering Department\\
	Universidad Tecnol\'{o}gica de Pereira, Pereira, Colombia\\
	{\tt\small gahol@utp.edu.co}
	\and
	Rachel Naegele \\
	United States Department of Agriculture, Agricultural Research Service,\\
	San Joaquin Valley Agricultural Sciences Center (USDA-ARS-SJVASC)\\
	Parlier, California, USA\\
	{\tt\small rachel.naegele@ars.usda.gov} 
}

\begin{document}
	\maketitle
	\thispagestyle{empty}
	\pagestyle{empty}

\begin{abstract}
	Images are used frequently in plant phenotyping to capture measurements. This chapter offers a repeatable method for capturing two-dimensional measurements of plant parts in field or laboratory settings using a variety of camera styles (cellular phone, DSLR), with the addition of a printed calibration pattern. The method is based on calibrating the camera using information available from the EXIF tags from the image, as well as visual information from the pattern. Code is provided to implement the method, as well as a dataset for testing.  We include steps to verify protocol correctness by imaging an artifact. The use of this protocol for two-dimensional plant phenotyping will allow data capture from different cameras and environments, with comparison on the same physical scale.  We abbreviate this method as CASS, for CAmera aS Scanner.
	
	
\end{abstract}

\section{Introduction}
\label{intro}

Images are used with increasing frequency in plant phenotyping for a variety of reasons.  One reason is the ability to remotely capture data without disturbing the plant material, while another is the promise of high-throughput phenotyping via image processing pipelines such as those enabled by PlantCV \cite{fahlgren_versatile_2015}.  

However, to acquire precise data suitable for measurements of two-di\-men\-sional objects, the prevailing method in the community is to use a flatbed scanner.  Shape analysis of leaves has used scanned images, for apple, grapevine, {\it Claytonia} L., and a mixture of species \cite{migicovsky_morphometrics_2018, klein_digital_2017, stoughton_next-generation_2018, li_topological_2018}\footnote{Note that not all of the data in \cite{li_topological_2018} is from a scanner.}.  Scanners have also been used to analyze the shape of pansy petals \cite{yoshioka_genetic_2006} and  {\it Vitis vinifera} L. seeds \cite{orru_morphological_2013}.  

Cameras have been used to phenotype a range of plant structures and sizes, such as cranberry fruit shape and size \cite{diaz-garcia_image-based_2018} and root system architecture \cite{das_digital_2015}. In both of these works, a disk of known diameter is added to the scene for scaling purposes.

\subsection{Camera calibration}

\begin{wrapfigure}{R}{4cm} 
	\centering
	\includegraphics[width=3cm]{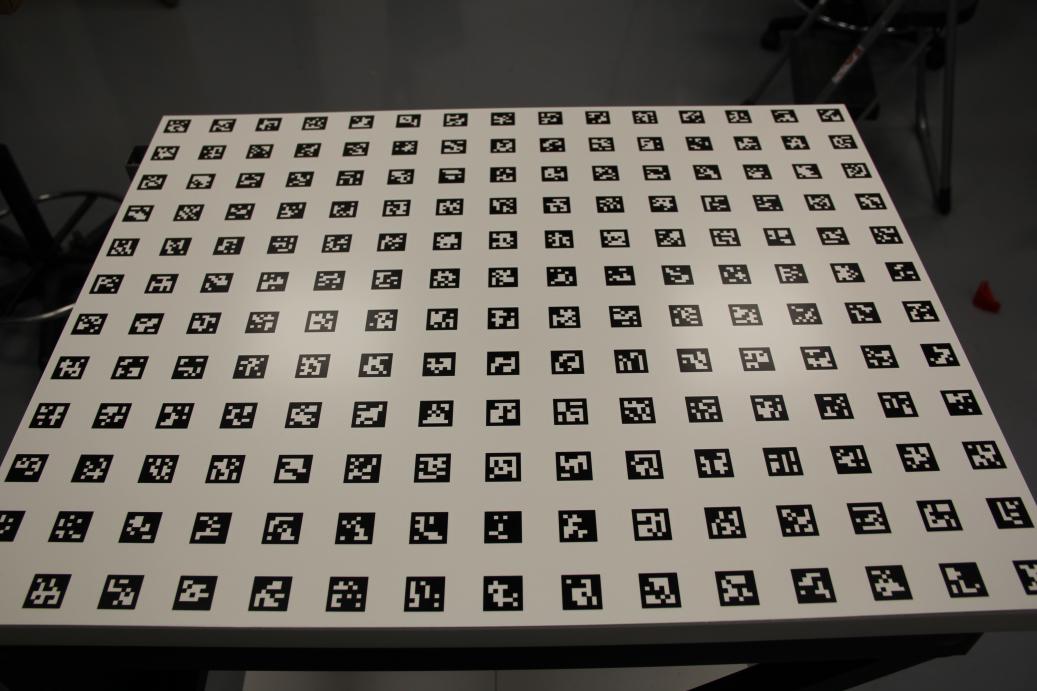} 
	\caption{An aruco calibration pattern. This particular example has been printed on aluminum, so it can be cleaned during experiments, which is convenient in plant research.}
	\label{fig:plain_pattern}
\end{wrapfigure}

The protocol in this paper transforms images acquired from a standard consumer camera such that measurements in pixels are representative of a planar scene.  What this means in more detail is that we have emulated a flatbed scanner with a consumer camera; angles between lines are preserved, as are distance ratios. Physical measurements can be recovered from image measurements by dividing by the number of pixels per millimeter, similar to flatbed scanners.  We abbreviate this method as CASS, for CAmera aS Scanner, using the tool from \cite{cook_acronym_2019}.

This method is needed because measurements of two-dimensional objects, when done in image space of camera-acquired images, are subject to diminished accuracy from physical perturbations. A small movement of the camera up or down will give the erroneous impression that an object is larger or smaller in terms of pixels.  Image pixels are also subject to radial distortion and projective geometry that allows three-dimensional objects to be viewed in a two-dimensional image.  In other words, $100$ pixels on one side of the image may not represent the same physical dimensions as $100$ pixels in another portion of the image.  

The method at the center of this protocol makes use of established camera calibration procedures to mitigate the problems of the preceding paragraph.  Camera calibration is the estimation of parameters that relate three coordinate systems: image, camera, and world, to each other.  This chapter does not have to the space to deal with this topic in depth, but Hartley and Zisserman \cite{hartley_multiple_2003} is a good text on this topic.  When camera calibration is completed, the coordinate systems have been defined relative to a standard, and the relationships of one coordinate system to another are known.  

Calibration patterns are used to define coordinate systems relative to a standard.  These may take many forms; in this work we use aruco patterns \cite{garrido-jurado_automatic_2014}; laid out in a grid, patterns define the X-Y plane of the world coordinate system as in Figure~\ref{fig:plain_pattern}.  The camera captures an image of the pattern to aid in defining the world coordinate system with respect to the image and camera coordinate systems.  

Usually, many views of the pattern are captured to solve an optimization problem to fully calibrate the camera \cite{zhang_flexible_2000}.  However, the Structure from Motion (SfM) community, \cite{snavely_photo_2006}, \cite{wu_towards_2013} began exploiting EXIF data, or Exchangeable image file format.  EXIF data is a type of meta data that is common in today's consumer cameras.  Within SfM, the camera's sensor size and some data from the EXIF file is used to generate an initial solution for some of the camera calibration parameters.  We have borrowed this practice for calibrating in the phenotyping context.

\subsection{Using CASS, camera as a scanner}

\begin{figure}    
	\centering
	\includegraphics[width=0.24\linewidth]{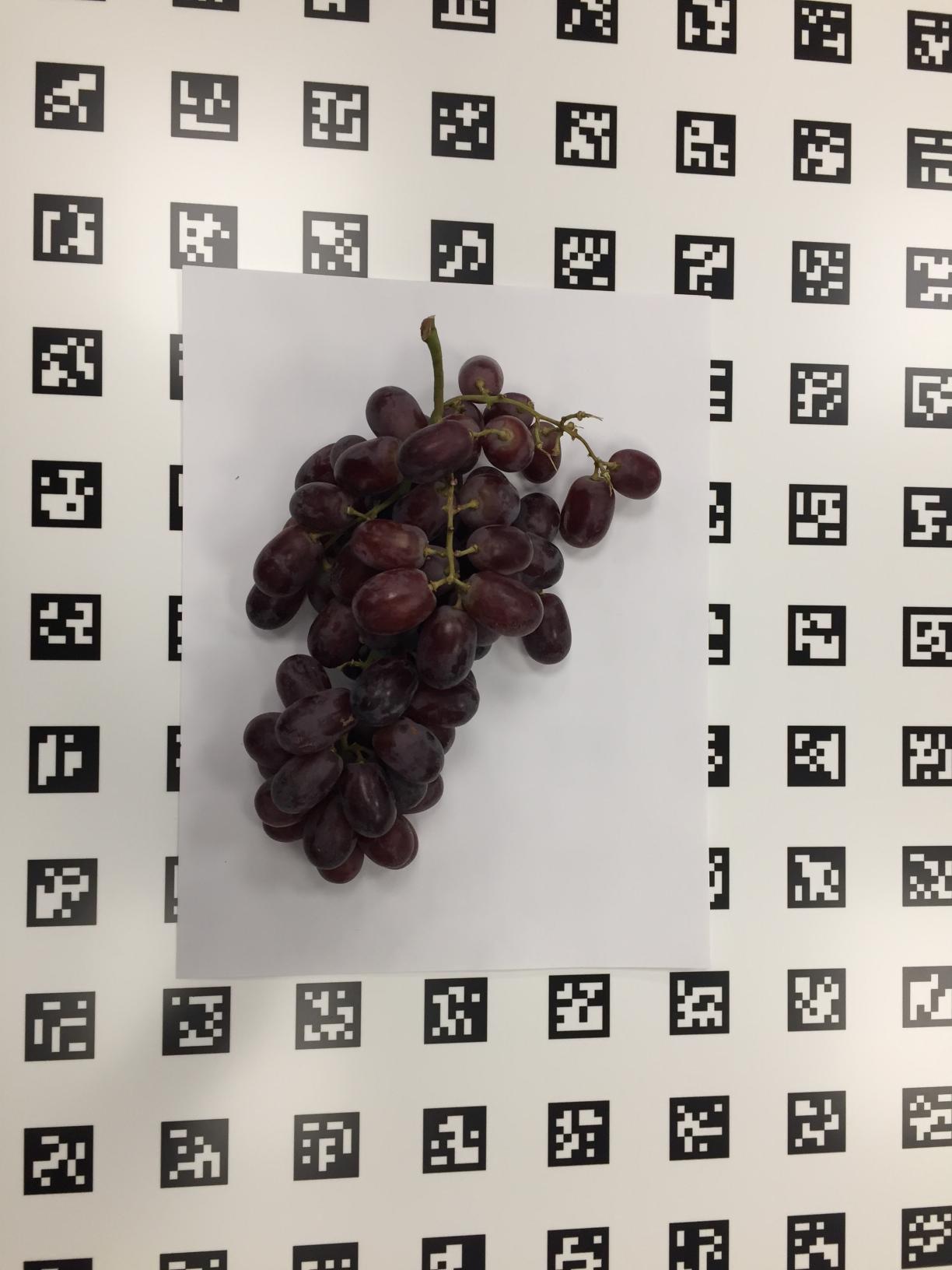} \includegraphics[width=0.24\linewidth]{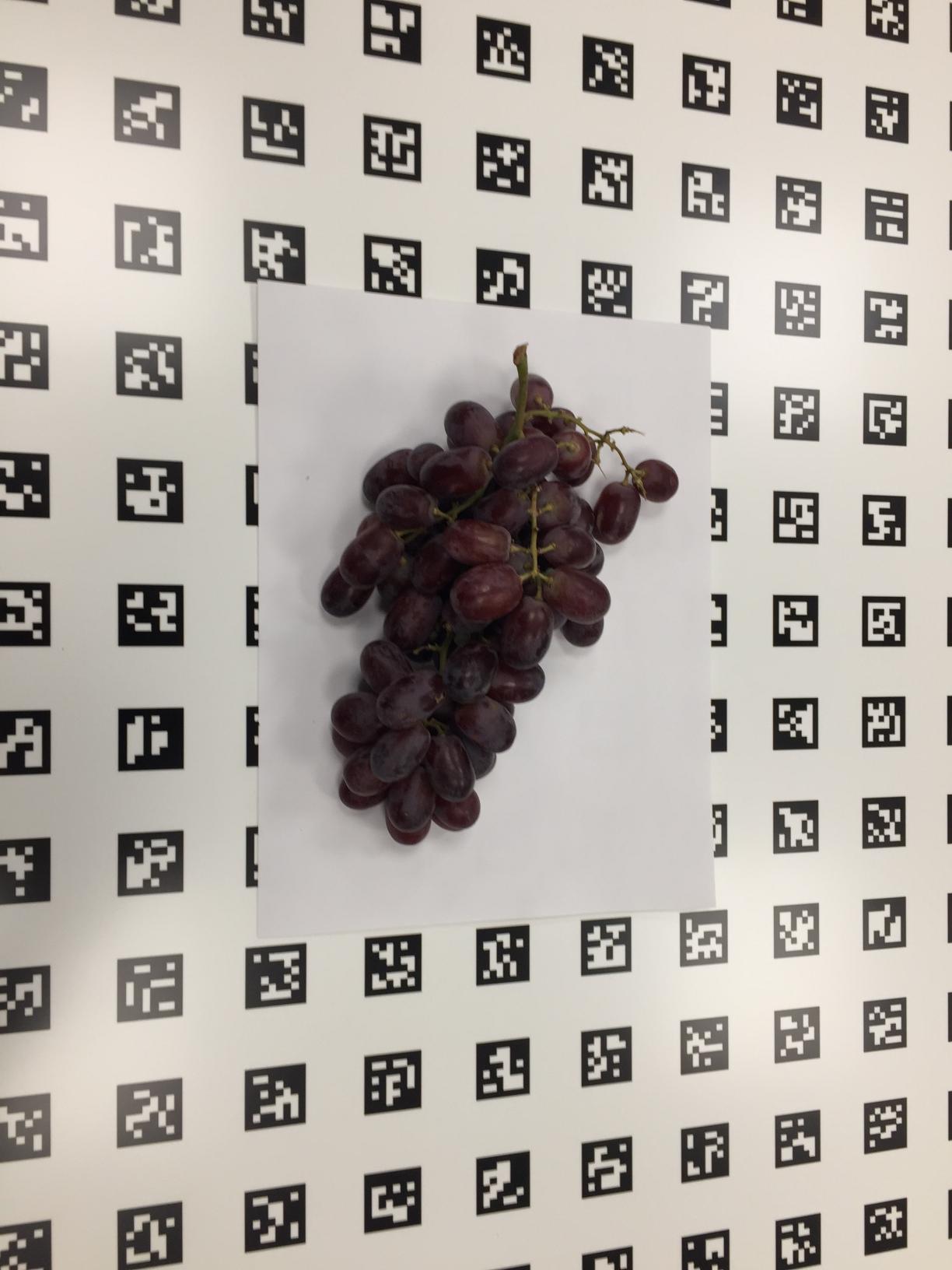}\includegraphics[width=0.24\linewidth]{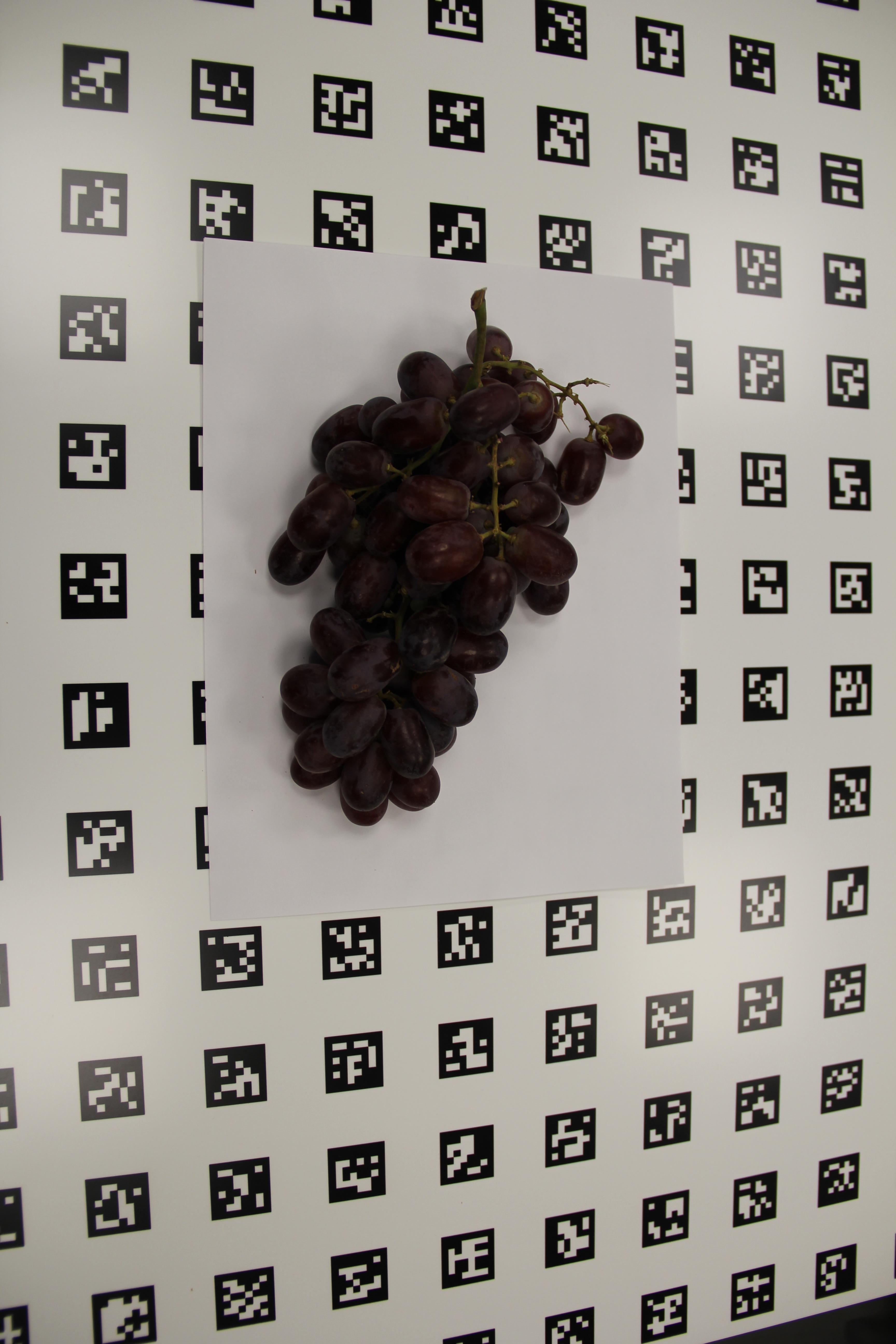} \includegraphics[width=0.24\linewidth]{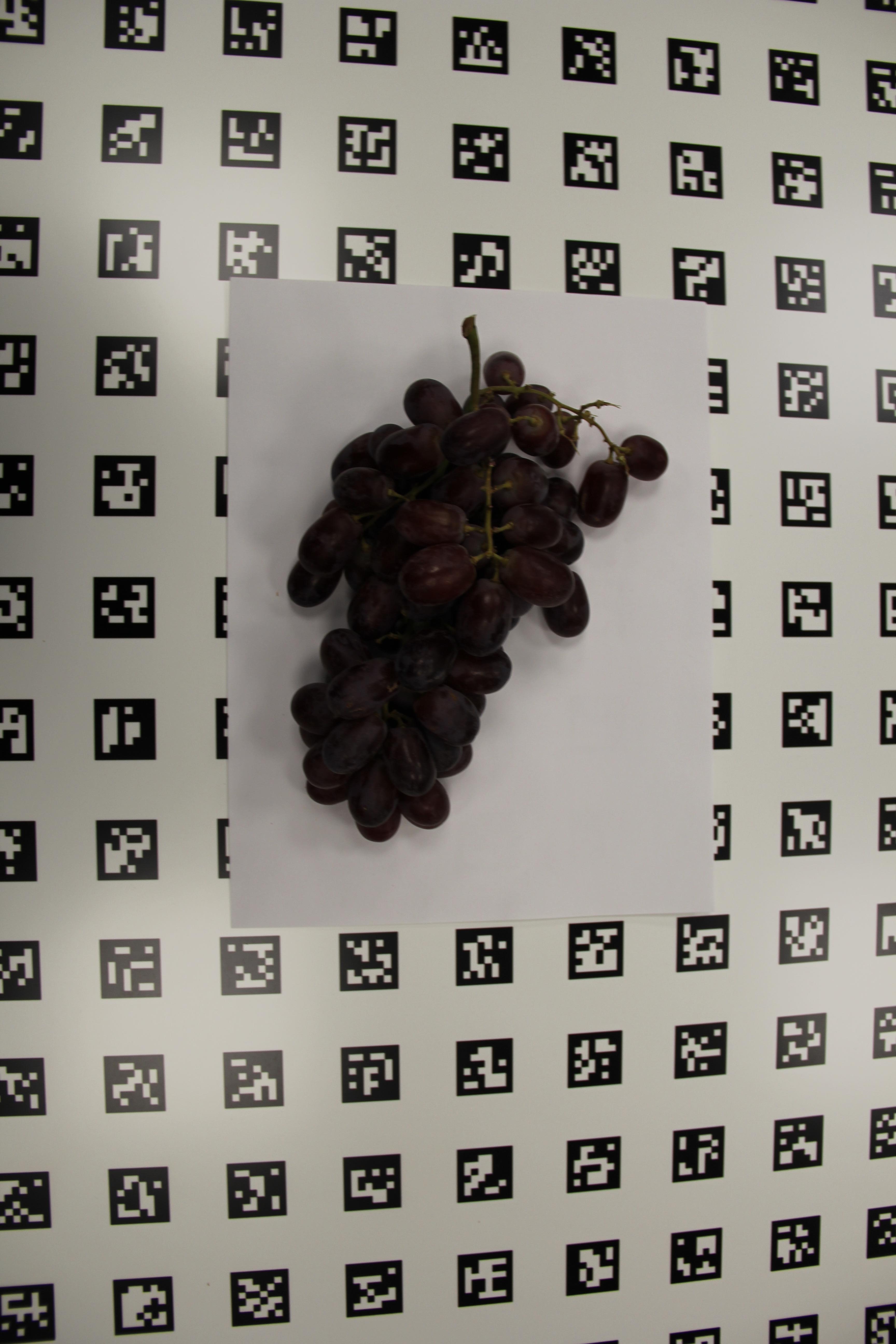}
	
	\includegraphics[width=0.24\linewidth]{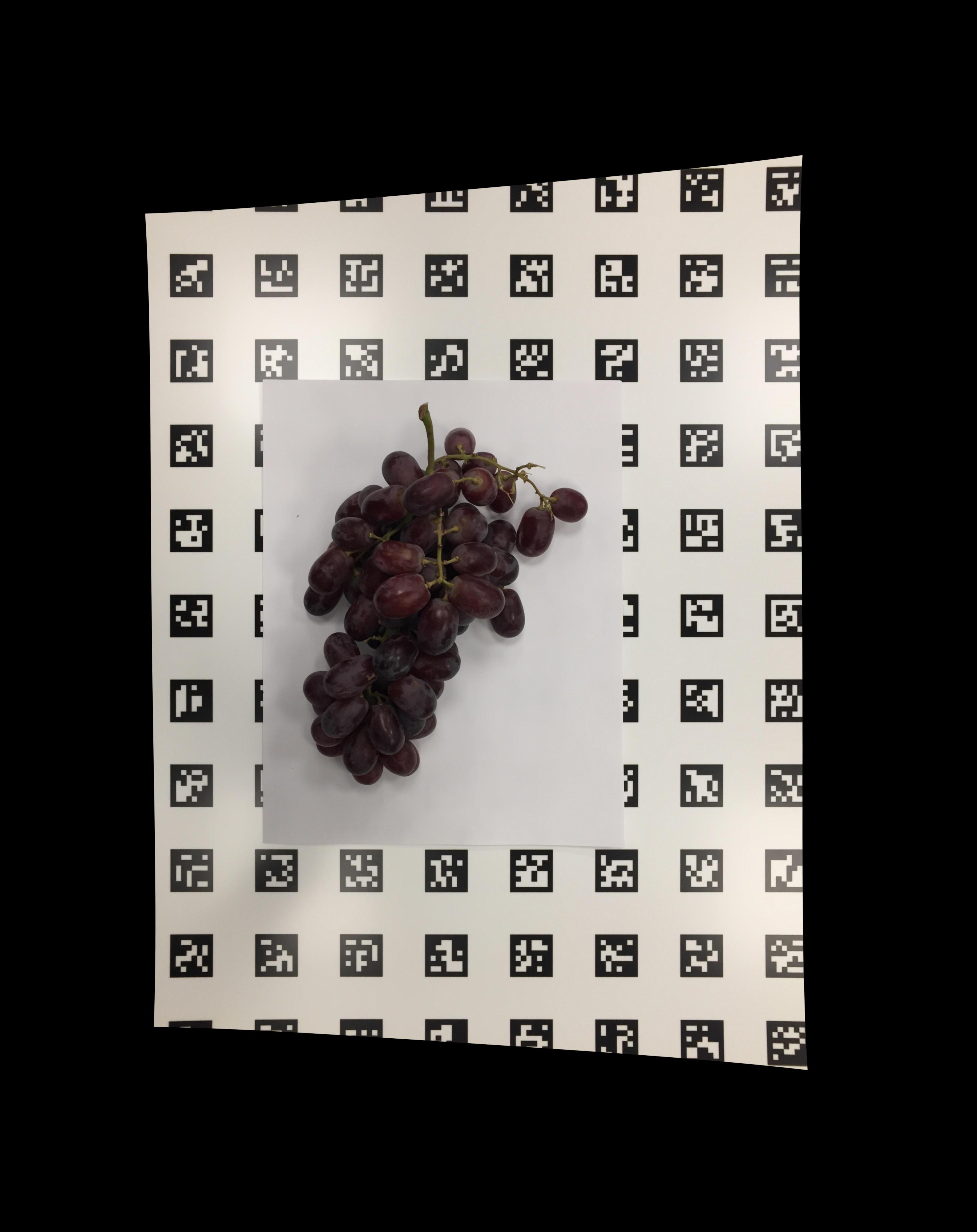} \includegraphics[width=0.24\linewidth]{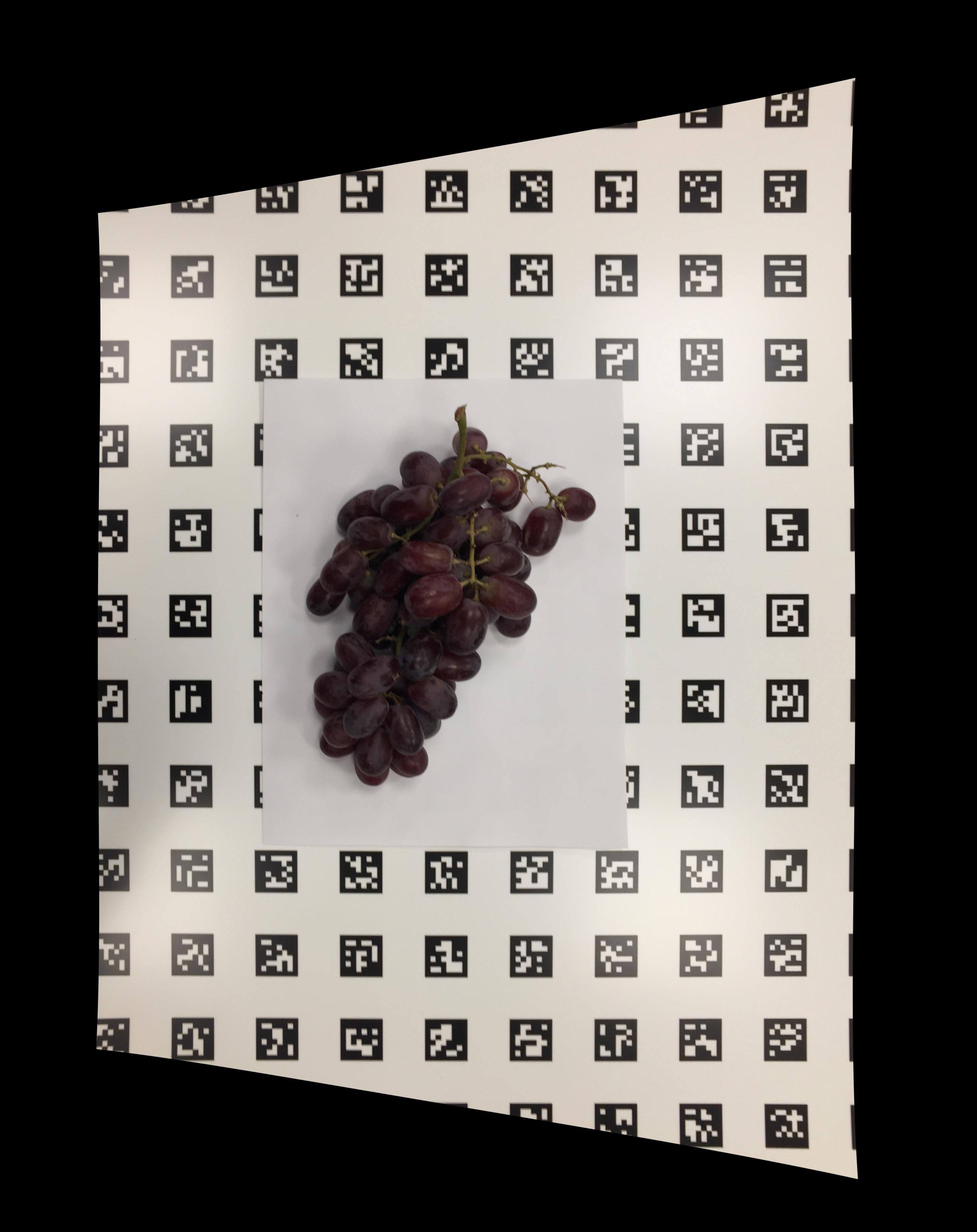}\includegraphics[width=0.24\linewidth]{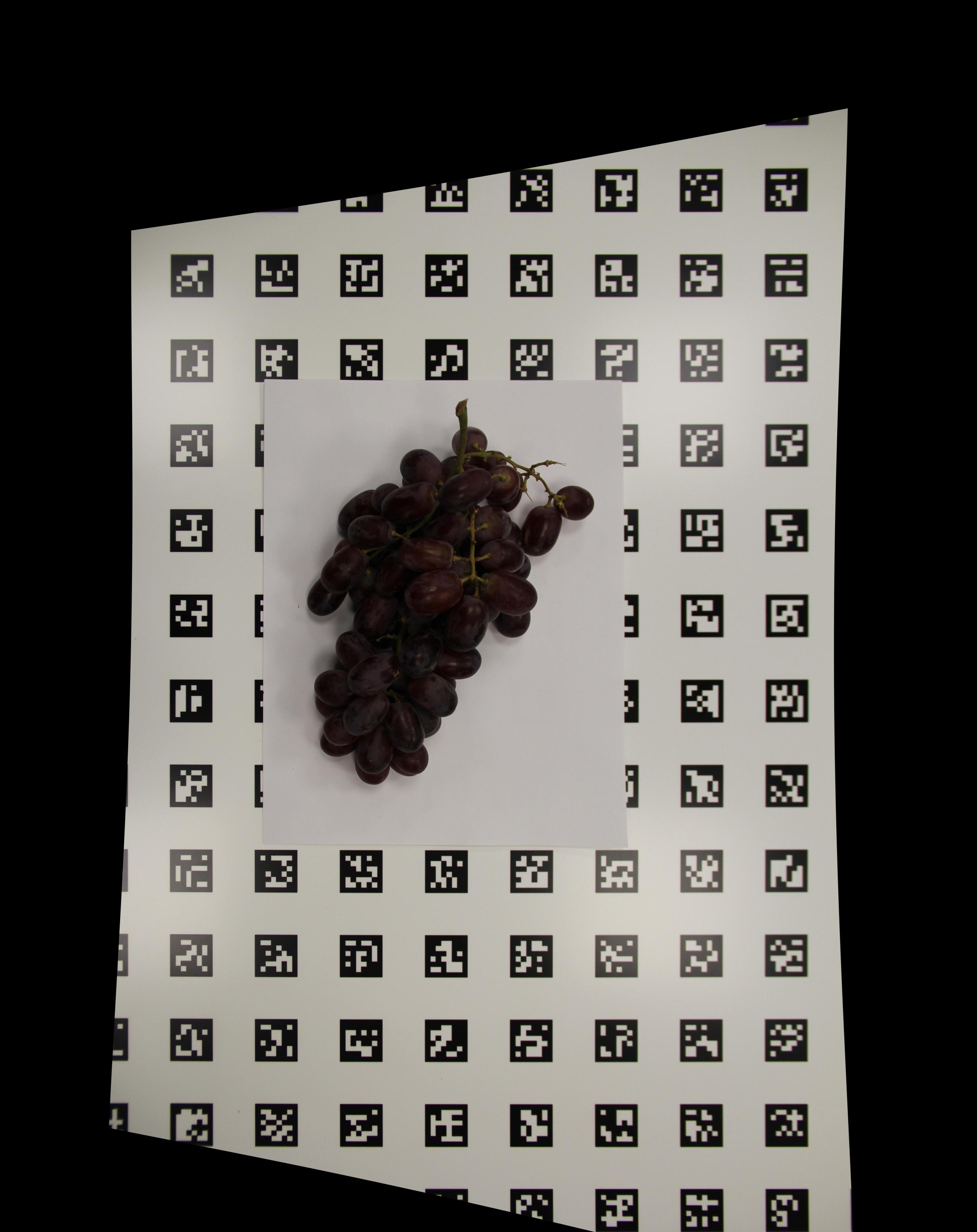} \includegraphics[width=0.24\linewidth]{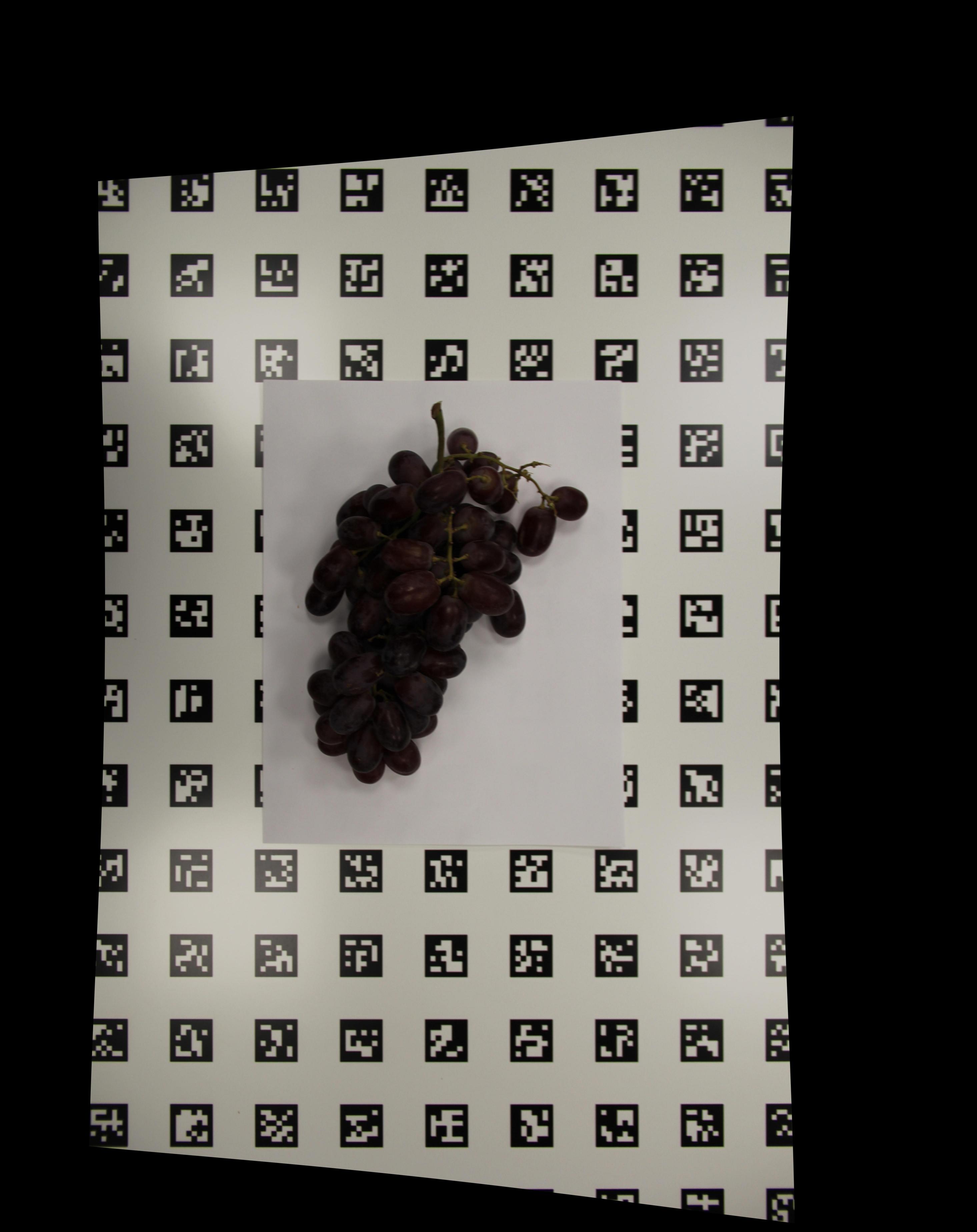}
	
	\caption{Example of a grape cluster.  This is a three-dimensional object, but we are interested in measuring aspects of the object where it meets the calibration pattern. Top row: input images of the same grape cluster, left two images are from an Apple iphone 6 (cellular phone camera), right two images are from a Canon EOS 60D {DSLR} camera.  Bottom row: results of applying the method for the image above, where every 10 pixels equal 1 millimeter. Full images are available in \cite{tabb_data_2020}.}
	\label{fig:overview_experiments}
\end{figure}

The original intent of this method, CASS, was to develop a high-throughput substitute for slow flatbed scanners. The steps in Section~\ref{sec:methods} will give details for the user.  A brief overview of the code is provided with this chapter: 1) calibrates the camera, per image, 2) computes the homography to transform the current image to the X-Y grid of the world coordinate system, and 3) warps the current image to match the world coordinate system's X-Y grid.

Figure~\ref{fig:overview_scanner} shows the input images and the output of CASS.  From the output images, users can apply their own computer vision techniques to identify the objects of interest.  Measurements in pixels can be transformed to physical units by dividing by the user-selected scaling factor. 

It is important to note a strong assumption when using this method, which is that the object is planar.  In practical terms, the user should either use objects that are roughly planar, or consider the footprint of the object on the calibration pattern plane.  CASS is not suitable for measuring objects that are non-planar, such as free-standing branches with the calibration pattern behind.  

To verify that the protocol has been performed correctly, we also include instructions for verifying that the measurements are correct by way of an artifact.

\section{Materials}
\label{sec:materials}

The materials needed are:
\begin{enumerate}
	\item{calibration pattern}
	\item{camera}
	\item{artifact}
	\item{code}
\end{enumerate}

\begin{figure}    
	\centering
	\includegraphics[width=0.33\linewidth]{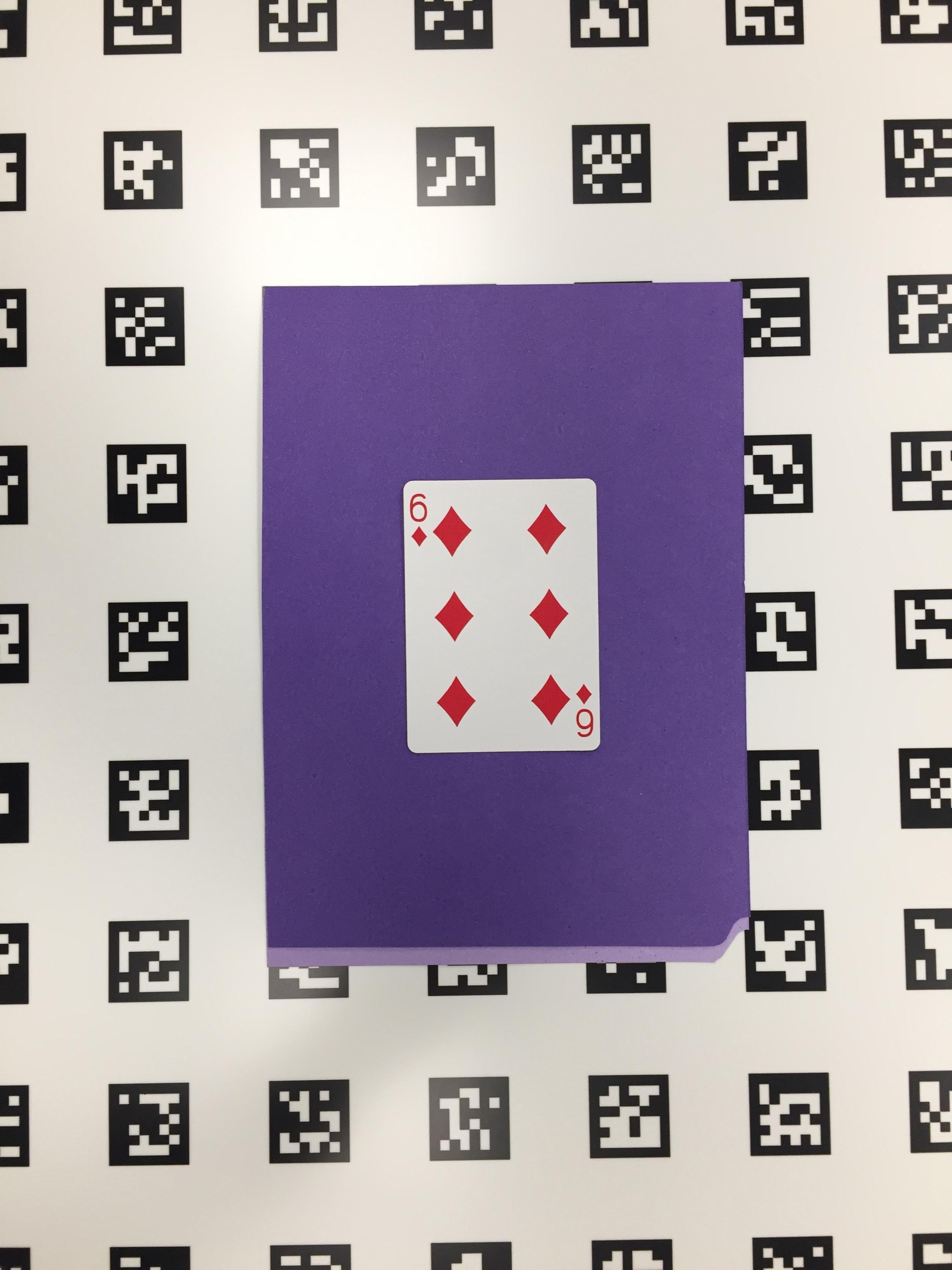} \includegraphics[width=0.33\linewidth]{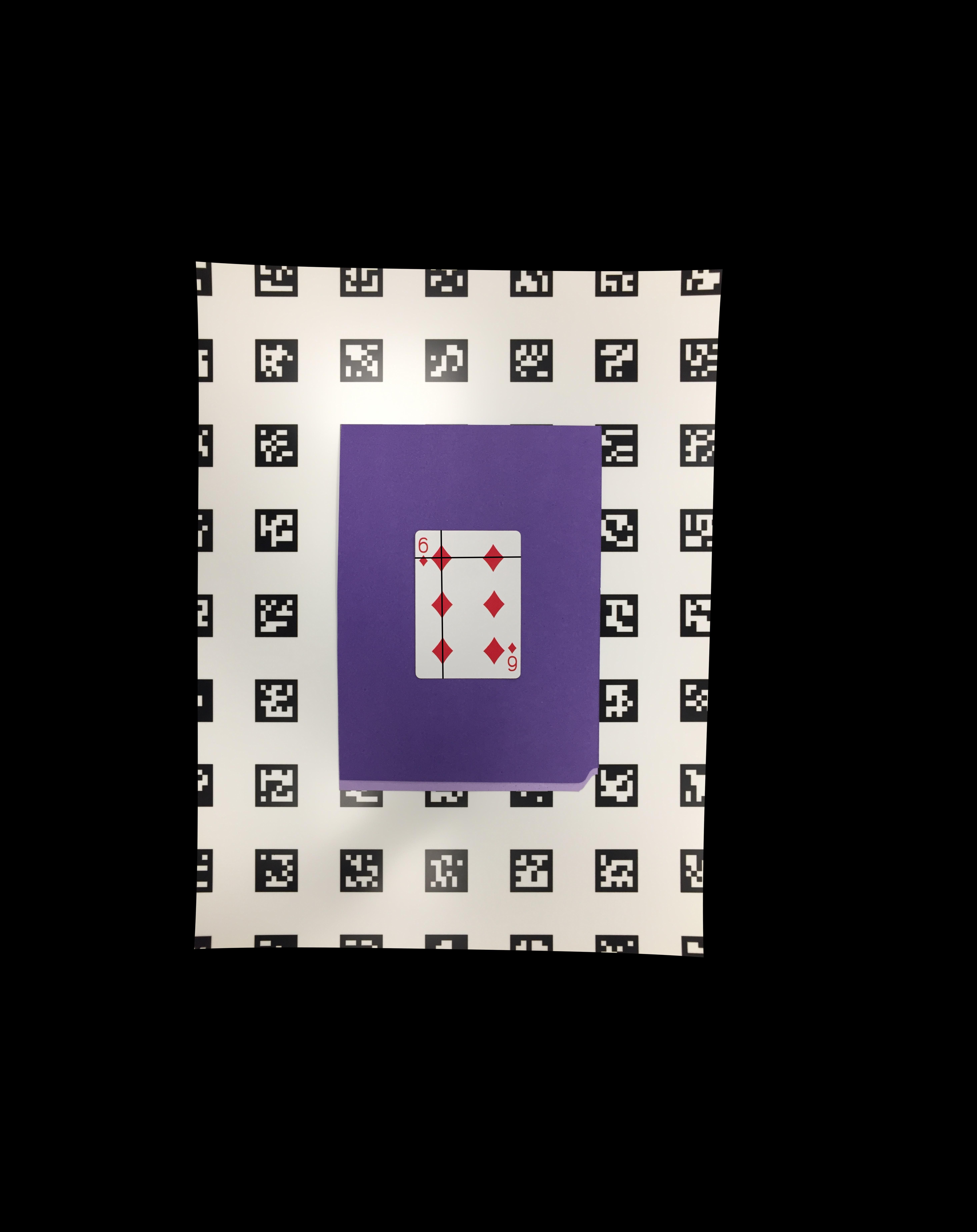}
	
	\caption{Left: Apple iphone6 camera images of a $2.5$ x $3.5$ inch ($63.5$ mm x $88.9$ mm) playing card. Right: results of applying the method for the image above, where every 10 pixels equal 1 millimeter.  Black lines indicate measurements of the card in pixels. The horizontal line was $635.02$ pixels, so is equivalent to $63.502$ mm as measured by this system.  The vertical line was $888.07$ pixels, which is equivalent to $88.807$ mm. }
	\label{fig:overview_scanner}
\end{figure}

The preparation of the calibration pattern is documented in~\ref{step1}.  The style of the camera is not specific to CASS, and should be chosen for the user's convenience.  This method relies on the extraction of EXIF tags, so the camera should write EXIF data.  At the time of this writing, this feature is common in consumer and cellular phone cameras.  An artifact of a known size is needed to check that the protocol has been implemented correctly.  In our example, we chose a playing card, as shown in Figure~\ref{fig:overview_scanner}.  A natural choice for an artifact may be a ruler.

The code and test datasets are provided in \cite{tabb_data_2020}.  Within \cite{tabb_data_2020}, are some example data and two programs: \\ {\tt aruco-pattern-write} and {\tt camera-as-scanner}. To prepare for the experiments, download the example data and install the code (C++ code as well as a Docker image are provided).  

\section{Methods}
\label{sec:methods}

\begin{enumerate}[leftmargin=*,label={Step \arabic*}, itemsep=1.2\baselineskip]
	\item{Prepare the aruco calibration pattern. The pattern should be printed such that $x$ and $y$ axes are equally scaled, and attached to a flat surface.  A pattern is provided in the \cite{tabb_data_2020} resource, as well as code for generating a new pattern via {\tt aruco-pattern-write} and instructions in its README.  Considerations when generating a new pattern are in~\ref{note1}.  The option of printing patterns on metal is discussed in~\ref{note2}. }\label{step1}
	
	\item{Arrange the object to be measured on top of the aruco pattern printout.  If segmentation of the object from the scene is desired using an image processing technique, we suggest placing a solid-colored paper or fabric in between the object and the pattern.  See~\ref{note3} for more details. }
	
	\item{Acquire images of the object, including at minimum a 1-layer border of aruco tags on all four sides of the image.  The image should generally be in focus, and acquired such that the camera body is parallel to the aruco pattern plane.  However, the alignment does not have to be exact.  See Figures~\ref{fig:overview_experiments} and~\ref{fig:overview_scanner} for examples.  If using a cell phone camera, do not zoom. Standard image formats are all acceptable, as long as EXIF tags are generated. }\label{acquire_images}
	
	\item{Acquire an image of an artifact (such as a ruler) of known size with the same protocol as in~\ref{acquire_images}.  We suggest that the artifact be rectangular in shape to allow for ease of measurement.}
	
	\item{Prepare the image and format information to run {\tt camera-as-scanner}.  This step assumes that the code has been installed according to its instructions, mentioned in Section~\ref{sec:materials}. }\label{step5}
	\begin{enumerate}[label={Step 5.\arabic*}]
		\item{The preparation instructions for running CASS for a group of images is given with the README of \cite{tabb_data_2020}. Create a test directory. }\label{readme}
		
		\item{Look up the camera's sensor size and convert to millimeters. This information may be found in the manufacturer's provided information that came with the camera, or can be found online.  Fill in the sensor size parameters in the appropriate file as indicated in~\ref{readme}.}
		
		\item{Measure one of the squares of the printed aruco calibration pattern, in millimeters.  Fill in the square length parameter of the appropriate file as as indicated in~\ref{readme}.}
		
		\item{Move the images of the objects and image of the artifact to a directory with the name {\it images} within the test directory.}
		
		\item{Determine the number of pixels per millimeter $np \in(0, \infty)$ for the transformed images, which will be an argument for running the code. The choice for $np$ depends on the size of the object, size of the calibration pattern, and how large one can tolerate the result image size.  Suppose the aruco calibration pattern print is $x$ mm $\times$ $y$ mm.  The result images will be $x * np$ pixels $\times$ $y * np$ pixels.  See~\ref{note4} for suggestions. In Figures~\ref{fig:overview_experiments} and~\ref{fig:overview_scanner}, $10$ was chosen.}\label{step55} 
	\end{enumerate}
	
	\item{Run the code {\tt camera-as-scanner} with three, and optionally four, arguments: the directory and the specified files and directory from~\ref{step5}, an output directory, and the number of pixels per millimeter $np$.  The arguments and format for them can be found by running the program with flag {\tt --help}. The optional fourth concerns whether intermediate results are written. }
	\item{Verify that the output is as expected, by inspecting the warped image corresponding to the artifact.  Measure the width of the artifact in an image manipulation program such as ImageJ, KolourPaint, the GIMP, Adobe Photoshop, etc.; its units will be pixels $w_p$.  Measure the width of the physical artifact in millimeters: $w_{mm}$.  The following should be true: $w_{mm} = \frac{w_p}{np}$. If not, then recheck the steps. The verification process was demonstrated with the playing card artifact in Figure~\ref{fig:overview_scanner}.}
\end{enumerate}

\section{Notes}
\label{sec:notes}

\begin{enumerate}[leftmargin=*,label={Note \arabic*}, itemsep=1.2\baselineskip]
	\item{ Note that the pattern can be scaled up or down to be suitable for the data acquisition context, such as the image provided in {\tt aruco-pattern-write} as an example.  It is not necessary for the camera to view the whole pattern. The patterns are black and white, so do not need to be printed in color.} \label{note1}
	
	\item{In our experiments, we have ordered prints of the patterns on aluminum. These have been convenient when working with fruit and plant material, because aluminum prints can be washed and cleaned.  It is important that the aruco patterns not become occluded with dirt or stains.} \label{note2}
	
	\item{Concerning segmentation of the object from the scene of aruco pattern and solid-colored fabric or paper, we suggest that the solid-colored fabric or paper be chosen such that it is a contrasting color compared to the target object. The fabric or paper should be cleaned or replaced if there are dirt or stains. The color of the fabric or paper, whatever color is chosen, will not interfere with the detection of the aruco tags.} \label{note3}
	
	\item{As $np$ increases, so will the image size. We suggest trying a range of sizes with a small number of images, such as $np = 5, 10, 20$, to get a sense of the resulting file size and resolution of features of interest.   
	}\label{note4}
\end{enumerate}

\section*{ACKNOWLEDGMENTS}
We acknowledge the support of USDA-NIFA Specialty Crops Research Initiative, VitisGen2 Project (award number 2017-51181-26829).

{\small
	\bibliographystyle{IEEEtran-dataset}
	\bibliography{refs}
}

\end{document}